%% file: main.tex
\documentclass[runningheads]{llncs}

\usepackage{eccv}

\usepackage{eccvabbrv}

\usepackage{graphicx}
\usepackage{booktabs}
\usepackage{multirow}
\usepackage{colortbl}
\usepackage{marvosym}

\usepackage[accsupp]{axessibility}  

\usepackage[pagebackref,breaklinks,colorlinks,citecolor=eccvblue]{hyperref}

\usepackage{orcidlink}

\begin{document}

\title{Atlas is Your Perfect Context: One-Shot Customization for Generalizable Foundational Medical Image Segmentation} 

\titlerunning{Atlas is Your Perfect Context}

\author{
Ziyu Zhang\inst{1}\thanks{Z. Zhang and Y. Yu---Equal contribution.}\orcidlink{0000-0003-1423-7036} \and
Yi Yu\inst{2}\textsuperscript{$\star$}\orcidlink{0000-0002-9841-4687} \and
Simeng Zhu\inst{2} \and
Ahmed H Aly\inst{2} \and
Yunhe Gao\inst{3} \and \\
Ning Gu\inst{1}\textsuperscript{(\Letter)}\orcidlink{0000-0003-0047-337X} \and
Yuan Xue\inst{2}\textsuperscript{(\Letter)}\orcidlink{0000-0002-5390-9037}}

\authorrunning{Z.~Zhang et al.}

\institute{Nanjing University \and
The Ohio State University \\
\email{Yuan.Xue@osumc.edu} \and
Stanford University\\
\url{https://github.com/yuyi1005/AtlasSegFM}}

\maketitle
\input{sec/0_abstract}    
\input{sec/1_intro}
\input{sec/2_related}
\input{sec/3_methods}
\input{sec/4_experiments}

\input{sec/5_conclusion}

\section*{Acknowledgements}
Ziyu Zhang was supported in part by the Natural Science Foundation of Jiangsu Province of China (BK20241202), the China Postdoctoral Science Foundation (2024M751394, GZC20240691), and the Jiangsu Funding Program for Excellent Postdoctoral Talent (2024ZB433).

%
%
\bibliographystyle{splncs04}
\bibliography{main}

\end{document}

%% file: sec/0_abstract.tex
\begin{abstract}
Accurate segmentation of anatomical structures in medical images is essential for diagnosis and treatment planning. While recent interactive segmentation foundation models enhance generalization through large-scale multimodal pretraining, they still depend on precise prompts and can fail in underrepresented clinical contexts (e.g., small organs-at-risk). We present AtlasSegFM, an atlas-guided framework that customizes off-the-shelf foundation models to new clinical contexts with a single annotated example. AtlasSegFM \textbf{1)} performs atlas-query registration to generate context-aware prompts, \textbf{2)} refines the segmentation with a frozen foundation model, and \textbf{3)} applies a lightweight adaptive fusion module to combine atlas priors with foundation-model inputs and predictions. Extensive experiments on six public and in-house datasets across radiotherapy and vascular scenarios show consistent gains, with the largest improvements on small and delicate structures. AtlasSegFM provides a lightweight, deployable solution for one-shot customization of segmentation foundation models in real-world clinical workflows.


\end{abstract}

%% file: sec/1_intro.tex
\section{Introduction}
\label{sec:intro}

Interactive segmentation has emerged as a promising approach in medical imaging, enabling precise and customizable delineation by incorporating user input \cite{kirillov2023ICCV_SAM, ravi2024_sam2, ma2024nc_medsam, ma2025_medsam2, zhang2026unlocking}. 
Unlike end-to-end methods, which may struggle to generalize across diverse anatomical structures and imaging modalities, interactive segmentation allows clinicians to guide the model so that results align with clinical expectations \cite{isensee2025_arxiv_nninteractive, wong2025iccv_multiverseg, zhang2025dcm}. 
In clinical practice, segmentation demands vary widely such as abdominal organ delineation, vascular segmentation in angiography, and organs-at-risk identification in radiotherapy, each with distinct anatomical complexity, task-specific requirements, and imaging protocols. Moreover, physicians may differ in how they define and prioritize structures, particularly for small yet clinically critical regions, making it essential for segmentation tools to accommodate individual judgment. The ability to perform segmentation on demand, customized to specific clinical contexts or physician instructions, is therefore crucial for rare conditions, patient-specific variations, or limited annotated data, underscoring the need for flexible, adaptive, and clinically intuitive solutions in medical image analysis.

\begin{figure}[t!]
\centering
\includegraphics[width=\linewidth]{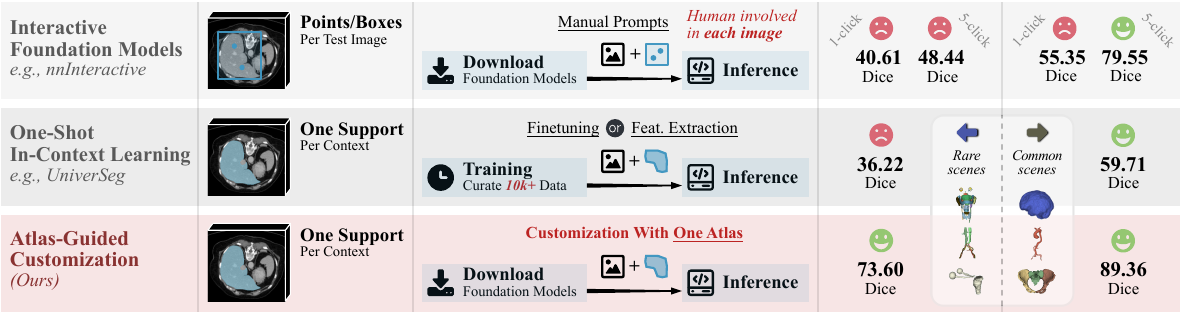}
\caption{Comparison of three segmentation paradigms. Interactive foundation models require per-image human prompts (points/boxes). One-shot in-context learning typically requires training on curated labeled datasets to learn cross-task prompting. Our AtlasSegFM adapts off-the-shelf foundation models to a target context using one annotated atlas. ``Rare-scene'' denotes contexts that are rare or not covered by the foundation model’s reported training distribution in our evaluation (e.g., organs-at-risk and lower-limb vessels; see Sec.~\ref{sec:experiments}).}
\label{fig:intro}
\vspace{0pt}
\end{figure}


Existing customized segmentation typically relies on two main approaches: interactive segmentation and In-Context Learning (ICL).
Interactive segmentation leverages user inputs, such as scribbles or clicks, to iteratively refine predictions and adapt to specific cases. 
While methods like MedSAM2 \cite{ma2025_medsam2}, nnInteractive \cite{isensee2025_arxiv_nninteractive}, and ScribblePrompt \cite{wong2024eccv_scribbleprompt} demonstrate its potential, they often require multiple interactions and rely heavily on the quality of user inputs.
In contrast, ICL enables segmentation by providing contextual examples or prompts during inference, without the need for interaction or user input. 
ICL methods such as UniverSeg \cite{butoi2023_ICCV_universeg} and Iris \cite{gao2025_CVPR_Iris} have shown the ability to generalize across various tasks and datasets by leveraging in-context examples, using reference examples or prompts to segment for specific scenarios.

As shown in Fig.~\ref{fig:intro}, interactive methods offer flexibility by leveraging user inputs (e.g., points or boxes) to iteratively refine predictions during inference, whereas one-shot in-context learning methods use a single support example per context and require no interaction at inference.
However, current ICL methods require large, carefully annotated datasets and training or extensive fine-tuning, making them both resource-intensive and time-consuming \cite{butoi2023_ICCV_universeg, wang2023_CVPR_Seggpt, rakic2024cvpr_tyche, gao2025_CVPR_Iris}. Meanwhile, the growing availability of segmentation foundation models suggests an alternative strategy: instead of training a new model for each task, we can customize strong pretrained models to each clinical context.
 


\begin{figure*}[t!]
\centering
\includegraphics[width=\linewidth]{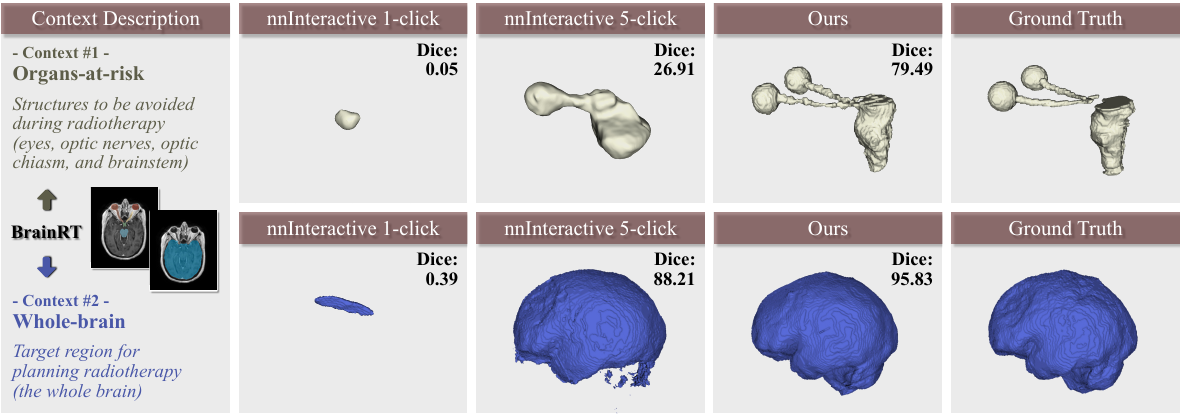}
\caption{Visual comparisons with recent interactive method (i.e., nnInteractive \cite{isensee2025_arxiv_nninteractive}) on the BrainRT dataset for radiotherapy. nnInteractive performs far below expectation in context \#1 (organs-at-risk), a rare scenario that is underrepresented in its training data, whereas our method remains robust in both contexts.}
\label{fig:vis-brain}
\vspace{0pt}
\end{figure*}

Moreover, interactive foundation models struggle when prompts are ambiguous or when the target structure is rare or small~\cite{ma2024nc_medsam, isensee2025_arxiv_nninteractive}.
As shown in Fig.~\ref{fig:vis-brain}, interactive methods perform well in common scenarios, such as whole brain segmentation, achieving a Dice of 88.21\% with 5-click interaction.
However, their effectiveness drops significantly in delicate structures, with the Dice falling to 26.91\% for organs-at-risk, where the targets are small, delicate structures embedded in complex context.
This gap reflects both the limitations of pre-training data coverage and the absence of explicit contextual information in spatial prompts, which impedes generalization to unseen categories and fine-grained anatomy.

To address the above challenges, we propose a novel pipeline that combines structural priors of classical atlas-based segmentation with the representational power of modern segmentation foundation models. 
Our method leverages atlas-based registration to avoid the training burden typically associated with ICL while preserving critical structural priors in a single support example.
The atlas mask serves as a context-aware prompt for interactive or non-interactive foundation models, providing more meaningful guidance and improving segmentation performance. 
This design enables efficient, contextually rich, and robust segmentation across diverse clinical scenarios.
The main contributions are as follows:
\begin{itemize}
    \item We investigate one-shot customization of segmentation foundation models for generalizable anatomical structure segmentation, where a frozen foundation model is adapted to a new clinical context using a single annotated atlas without requiring a dedicated training set.

    \item We present AtlasSegFM, a simple yet effective pipeline that combines the global anatomical consistency of atlas registration with the local refinement capability of segmentation foundation models. A lightweight adaptive fusion module integrates their complementary strengths at test time.

    \item Evaluated on multiple public and in-house datasets spanning organs-at-risk, pelvic anatomy, and vascular trees, AtlasSegFM consistently improves segmentation accuracy and boundary quality, particularly in underrepresented anatomical contexts where existing methods often fail, demonstrating strong out-of-distribution generalization.
\end{itemize}

%% file: sec/2_related.tex
\section{Related Work}
\label{sec:related}
\subsection{Medical Image Segmentation}
Medical image segmentation has long been an active research area, with deep learning emerging as the dominant approach. Early deep learning methods relied on supervised learning, leading to the development of numerous architectures, including U-Net variants \cite{ronneberger2015_UNet, zhou2019TMI_unet++}, Transformer-based models \cite{gao2021miccai_UTNet, wang2022ICASSP_mtunet, zhou2023tmi_nnformer}, and frameworks based on Mamba \cite{xing2024miccai_segmamba, liu2024tmi_swinUMamba, ruan2024acm_vmunet}. 
These methods have demonstrated remarkable performance across a wide range of segmentation tasks, particularly when sufficient annotated data is available.
By incorporating an automated pipeline, nnU-Net \cite{isensee2021_NM_nnunet} optimizes preprocessing, architecture, and hyperparameters, providing a robust and adaptable solution.
Its comprehensive design has set a standard for supervised learning methods in the field.

However, these methods face significant challenges when dealing with out-of-distribution data or new segmentation tasks \cite{zhang2024CVPR_mapseg}. 
Their performance deteriorates in scenarios that deviate from the training domain, and adapting to new or specific tasks often requires retraining the model from scratch. 
This process is heavily reliant on annotated data and substantial computational resources, making it impractical for deployment in highly customized clinical tasks.

\subsection{Segmentation Foundation Models}\label{sec:related_fm}
Foundation models have shown remarkable success in medical image segmentation \cite{kirillov2023ICCV_SAM,ravi2024_sam2}.
Some are task-specific, such as vesselFM \cite{wittmann_2025CVPR_vesselfm}, a general-purpose zero-shot 3D vascular segmentation model with strong generalization capability.
To enable user-customized segmentation, another line of research has explored interactive approaches, such as MedSAM \cite{ma2024nc_medsam, ma2025_medsam2} and nnInteractive \cite{isensee2025_arxiv_nninteractive}.
By incorporating user prompts (e.g., clicks and bounding boxes), these methods simplify the annotation and achieve strong performance with minimal user interaction \cite{wong2024eccv_scribbleprompt, gong2024mia_3dsam-adapter, wang2025tnnls_SAMMED3D, du2024segvol, guo2025towards}.

While these models leverage powerful pretraining on large datasets to generalize across tasks, they still face two challenges:
\textbf{1)} Prompts are often ambiguous, and the models infer user intent based on prior training data. Consequently, in rare contexts (e.g., organs-at-risk in Fig.~\ref{fig:vis-brain}), their performance falls short of expectations.
\textbf{2)} They require repeated user interactions for each image, making them inefficient for large-scale segmentation tasks or routine clinical workflows, where many patients follow similar procedures.
In-context learning is a possible solution to these challenges.

\subsection{In-Context Learning Methods}
In-context learning allows models to generalize to unseen tasks by leveraging a context set of labeled image-segmentation pairs at inference time \cite{hu2024midl_iclsam, wu2024CVPR_oneprompt, xie2025tmi_eicseg,zhao2025pr_segmic, wong2025iccv_multiverseg}.
Few-shot segmentation methods \cite{butoi2023_ICCV_universeg, lv2023aaai_styleseg}, primarily designed to reduce labeling effort, were explored first. However, most of these methods rely on small models and limited data, and some \cite{ouyang2022_TMI_SSLALP, lin2023_MICCAI_CATNET, cheng2024_TMI_GMRD, tang2025_MedIA_DSPNet} are further fine-tuned on the target domain (referred to as cross-domain ICL). As a result, they lack the generality needed to perform arbitrary segmentation like foundation models. Therefore, some ICL studies train on larger datasets, including UniverSeg \cite{butoi2023_ICCV_universeg} and Iris \cite{gao2025_CVPR_Iris}.

With the growing availability of pretrained foundation models (see Sec.~\ref{sec:related_fm}), a key limitation of these ICL approaches has become evident---they must be trained from scratch or fine-tuned on specific datasets, preventing them from leveraging the continuously improving foundation models. 
To address this, some studies have incorporated foundation models into ICL \cite{xu2025ICML_unlocking}. Their application to the medical domain is still at an early stage \cite{liu2025ijcv_segment, mao2025ICCV_onePolyp, zhu2025_MICCAI_maup}, with either unavailable code or limited performance. Moreover, existing ICL methods are primarily designed and evaluated on 2D images, limiting their applicability to 3D segmentation. How to fully leverage the capabilities of foundation models in ICL remains largely unexplored in the literature. 

\subsection{Atlas-Based Segmentation}
%
Atlas-based methods represent a classical approach that does not require annotations or learning \cite{iglesias2015multi}. 
These methods treat segmentation as a registration problem, where the atlas is aligned to unlabeled data using optimization algorithms. 
Through this alignment, the structural information from the atlas can be transferred to the target image, enabling segmentation without direct supervision.
Historically, atlas-based approaches iteratively minimize a cost function measuring the similarity between atlas and query \cite{wang2012multi}. More recently, some deep-learning-based registration methods have been proposed \cite{balakrishnan2019TMI_voxelmorph, hoffmann2021tmi_synthmorph, wang2024tmi_rdp}.
By leveraging neural networks, these approaches efficiently learn the mappings from atlas to query, significantly reducing computational overhead while improving accuracy.

%% file: sec/3_methods.tex
\section{Methods}
\label{sec:methods}

Medical image segmentation aims to assign a label to each pixel (or voxel) of an input image $X \in \mathbb{R}^{H \times W \times D}$ (e.g., CT/MRI scans) based on its underlying anatomical structure.
Given a segmentation model $f(\cdot)$, the goal is to predict a label map $ Y \in \mathbb{R}^{H \times W \times D} $, where each voxel $ y_i \in Y $ corresponds to a predefined anatomical or pathological category, as $Y = f(X)$.

In clinical practice, segmentation often needs to be \textit{Customized} for specific tasks, such as segmenting rare structures or adapting to new imaging modalities. 
Inspired by ICL, where a support set is used to provide domain-specific guidance, a contextualized segmentation task can be expressed as:
\begin{align}
    Y_{\mathit{query}} = f(X_{\mathit{query}}, \mathcal{C})
\end{align}
where the context support $\mathcal{C}$ provides task- and anatomy-specific guidance. In AtlasSegFM, $ \mathcal{C} = (X_{\mathit{support}}, Y_{\mathit{support}}) $ is a a single annotated atlas (one support pair) for the target structure (or a target class in multi-class datasets) and $ X_{\mathit{query}} $ is the query image. 

\begin{figure*}[t!]
\centering
\includegraphics[width=\linewidth]{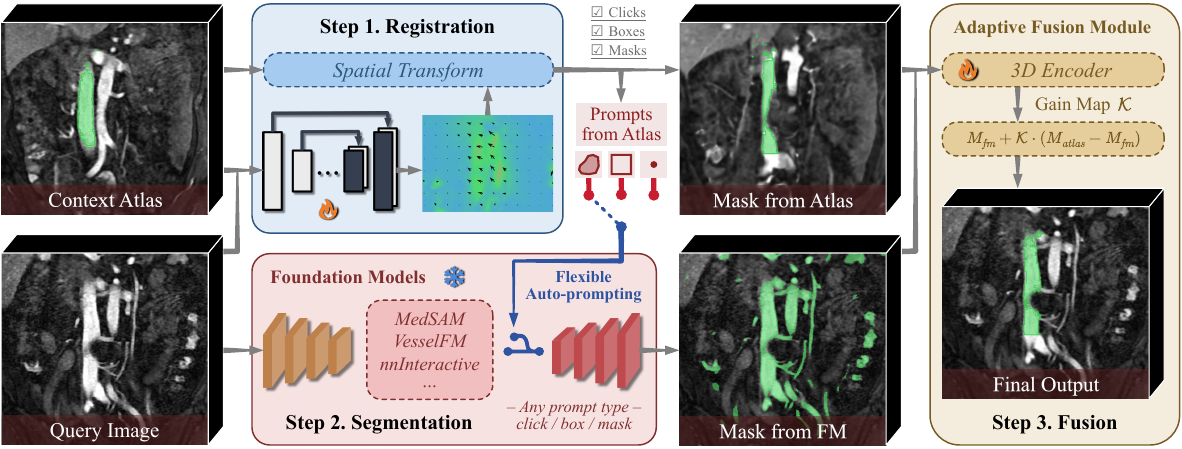}
\caption{Pipeline including three steps: \textbf{1)} Registration between query and support to obtain the mask (see Sec.~\ref{sec:registion}), \textbf{2)} Prompting the foundation model based on the mask from atlas (see Sec.~\ref{sec:foundation_model}), and \textbf{3)} Fusion of the two masks to obtain the final result (see Sec.~\ref{sec:fusion}). Our model uses an inference-only design, where the ``fire'' denotes test-time adaptation and the ``snow'' remains frozen.}
\label{fig:main}
\vspace{0pt}
\end{figure*}

Existing ICL methods rely on feature similarity between $ X_{\mathit{support}} $ and $ X_{\mathit{query}} $, which either requires extensive training and large-scale labeled data (e.g. UniverSeg\cite{butoi2023_ICCV_universeg}, Iris\cite{gao2025_CVPR_Iris}) or depend on large pretrained encoders such as DINO \cite{oquab2023_dinov2} for feature extraction.
As shown in Fig.~\ref{fig:main}, we propose an atlas-guided, in-context segmentation framework that combines the strengths of classical atlas-based registration and advanced foundation models.
AtlasSegFM consists of three steps: \textbf{1)} Atlas-based registration to provide context-aware prompts, \textbf{2)} Segmentation refinement using foundation models with these prompts, and \textbf{3)} Fusion module to integrate predictions from both sources. 

\subsection{Registration and Prompt Generation}\label{sec:registion}
To generate task-specific prompts for foundation models, we adopt an atlas-based registration approach as the initialization step for in-context learning with two main advantages:
\textbf{1)} atlas-based registration does not require training or labeled datasets, making it lightweight and easy to deploy in clinical workflows,
and \textbf{2)} it inherently leverages structural priors, which are particularly valuable for medical image segmentation tasks where anatomical consistency is critical.

Traditional atlas-based segmentation relies on iterative optimization to align the atlas to the query image, which is computationally expensive and time-consuming. To overcome this, we develop a test-time registration network derived from RDP \cite{wang2024tmi_rdp}, utilizing its network architecture and optimization objectives with adaptations for test-time optimization. 
Formally, given a context atlas $(X_\mathit{atlas}, Y_\mathit{atlas})$ and a query image $X_\mathit{query}$, we estimate the spatial transformation $T$ by optimizing:
\begin{align}
    T^\star = \arg\min_T \mathcal{L}_{\mathit{reg}}(T(X_{\mathit{atlas}}), X_{\mathit{query}})
\end{align}
where $\mathcal{L}_{\mathit{reg}}$ is a similarity-based registration loss, such as normalized cross-correlation or mutual information.
The registered atlas $T^\star (X_{\mathit{atlas}})$ and its corresponding label $T^\star (Y_{\mathit{atlas}})$ yield a coarse segmentation mask $M_\mathit{atlas}$, which serves as a structural prior.

Original RDP optimizes registration by training on large datasets to learn network parameters. In AtlasSegFM, we repurpose it as a test-time optimization framework, where the network parameters are optimized on-the-fly for each specific atlas-query pair during inference. 
In addition, we perform rigid and affine pre-registration prior to network optimization, ensuring coarse alignment and enhancing final registration accuracy (as validated in the ablation Table~\ref{tab:ablation}).

After generating the coarse segmentation mask $M_\mathit{atlas}$ from the registered atlas, we adapt it as the prompt context for foundation models.
Specifically, our framework can provide three different types of prompts:
\begin{itemize}
    \item Click: The centroid of the largest connected region within $M_\mathit{atlas}$ is extracted as the click prompt $P_\mathit{click}$.
    \item Box: The minimum circumscribed bounding box that encloses $M_\mathit{atlas}$ at the middle slice along $z$-axis is computed as the box prompt $P_\mathit{box}$.
    \item Mask: The segmentation mask $M_\mathit{atlas}$ is directly used as the mask prompt $P_\mathit{mask}$ for foundation models that support mask prompting.
\end{itemize}
These atlas-derived prompts inject explicit anatomical context (location, extent, and plausible shape) into the foundation model. Additional quantitative evidence comparing prompt types is reported in Sec.~\ref{sec:ablation}.


Following the inference protocols of nnInteractive \cite{isensee2025_arxiv_nninteractive} and MedSAM2 \cite{ma2025_medsam2}, point and box prompts are provided on one slice. 
Specifically, we derive the 2D bounding box from the middle slice of $M_\mathit{atlas}$.
nnInteractive utilizes this input to directly infer the 3D volume, whereas MedSAM2 initiates segmentation at the middle slice and propagates the prediction bidirectionally through the volume.

\subsection{Foundation Model Inference with Auto-Prompts}\label{sec:foundation_model}
To seamlessly integrate different foundation models, our pipeline is designed to support both interactive and non-interactive foundation models. For promptable models, prompts generated in the previous step, denoted as $\mathit{P}_i$ with $i \in \{\mathit{click}, \mathit{box}, \mathit{mask}\}$, are used to guide the segmentation refinement. The foundation model can select any supported prompt type (i.e., click, box, or mask) according to its interface, enabling flexible integration across models with different prompting mechanisms. For foundation models that do not support prompting (i.e. vesselFM \cite{wittmann_2025CVPR_vesselfm}), the same pipeline simply feeds the input images directly to the foundation model without prompts. In our experiments with nnInteractive and MedSAM2, we use atlas-derived mask prompts by default, and analyze click/box/mask prompting in Sec.~\ref{sec:ablation}.

This unified auto-prompting design allows our pipeline to flexibly adapt to different foundation models and scenarios, leveraging prompting when available while remaining compatible with models that operate without prompts.
The foundation model output can be expressed as:
\begin{align}
    M_\mathit{fm} = f_\mathit{FM}(X_\mathit{query},\mathit{P}_i)
\end{align}
where $f_\mathit{FM}$represents the foundation model and $\mathit{P}_i$ is the prompt from atlas.

\subsection{Atlas-FM Adaptive Fusion}\label{sec:fusion}
Although the foundation model provides fine-grained segmentation $M_\mathit{fm}$, it may occasionally fail in ambiguous regions or under challenging scenarios.
On the other hand, the atlas-based mask $M_\mathit{atlas}$ offers robust structural priors but generally lacks fine detail.
To address the limitations of both methods, we design a fusion module to dynamically integrate the predictions from the atlas and the foundation model, leveraging their complementary strengths.

The final mask $M_\mathit{final}$ is computed by combining $M_\mathit{atlas}$ and $M_\mathit{fm}$ through a voxel-wise reliability-gated fusion: 
\begin{align}
    M_\mathit{final} = M_\mathit{fm} + \mathcal{K} \cdot (M_\mathit{atlas} - M_\mathit{fm})
\end{align}
where $ \mathcal{K}$ is a spatial gain map that selects which source to trust at each voxel.
A larger $\mathcal{K}$ indicates higher local reliability of the atlas prediction relative to the FM prediction, whereas a smaller $\mathcal{K}$ makes the output rely more on the FM.

Rather than using a fixed or global fusion weight, we estimate $\mathcal{K}$ using a lightweight adaptive reliability estimator.
The estimator takes the query image $X_{\mathit{query}}$ and the two predictions $M_{\mathit{atlas}}$ and $M_{\mathit{fm}}$ as inputs, and predicts a voxel-wise reliability gate as: 
\begin{align}
     \mathcal{K} = \sigma (g([X_{\mathit{query}}, M_\mathit{atlas}, M_\mathit{fm}]))
\end{align}
where $\sigma(\cdot)$ is the sigmoid activation function to ensure $ \mathcal{K} \in[0,1]$.
In practice, $g(\cdot)$ first derives a feature volume from the normalized query image, the logit representations of $M_{\mathit{fm}}$ and $M_{\mathit{atlas}}$, their signed logit difference, the voxel-wise disagreement between the two soft masks, and the entropy maps of both predictions.
These cues encode local anatomical context, mask conflict, and uncertainty, enabling the network to estimate which source is more reliable at each voxel during test-time adaptation.
The feature volume is then processed by a lightweight 3D encoder followed by a sigmoid activated gate.
The resulting gate adaptively controls the fusion between the foundation-model prediction and the atlas prediction.
The estimator parameters are optimized at test time using a cycle-transformation loss, as detailed in the implementation details.

%% file: sec/4_experiments.tex
\section{Experiments}
\label{sec:experiments}

\textbf{Datasets.}
We evaluate our proposed method on six datasets, covering commonly used few-shot medical image segmentation tasks and clinically significant challenges.
To ensure a fair evaluation of generalization, all six datasets (public and in-house) are excluded from the nnInteractive training corpus, and we use fully held-out splits for all experiments.
\begin{itemize}
    \item \textbf{HaN-Seg} \cite{podobnik2023mp_hanseg}: A head and neck dataset contains 42 CT scans annotated for organs-at-risk. It serves as a standard benchmark for radiotherapy planning, presenting significant challenges in segmenting densely packed and low-contrast anatomical structures within the complex head and neck region.
    \item \textbf{SegRap} \cite{luo2025MIA_segrap2023}: Targeting segmentation for radiotherapy in nasopharyngeal carcinoma, this dataset contains 30 CT scans annotated for organs-at-risk, many of which are distorted or affected by nearby tumors.
    \item \textbf{Pengwin} \cite{liu2023miccai_pengwin}: A pelvic bone segmentation dataset containing 100 CT scans annotated for various pelvic anatomical structures. It presents unique challenges in delineating complex bone geometries, articulating joints, and varying bone densities across different patients.
    \item \textbf{AVT} \cite{radl2022Data_avt}: Aortic vessel tree dataset contains 56 CTA scans for vascular segmentation with segmentation masks of the aorta and its branches.
    \item \textbf{Fe-MRA}: An in-house dataset of 50 ferumoxytol-enhanced MRA scans of the lower extremities, with arterial and venous structures annotated into 12 fine-grained categories such as great saphenous veins and femoral arteries. It targets lower-limb vasculature in clinical scenarios including diagnosis and treatment planning for varicose veins and arterial thrombosis.
    \item \textbf{BrainRT}: An in-house dataset of 60 CT/MR scans for brain tumor radiotherapy planning, where radiologists delineated organs on CT and labels were registered to MR. It covers two crucial segmentation contexts: whole brain and organs-at-risk (eyes, optic nerves, optic chiasm, and brainstem), whose delineation is essential to minimize radiation-induced damage.
\end{itemize}
 
\noindent\textbf{Protocol and metrics.} Following \cite{ouyang2022_TMI_SSLALP}, we adopt a 5-fold cross-validation protocol, where one sample from the remaining validation set is used as the context support for testing the query images in each fold. For datasets with multiple categories, each category is treated as a segmentation context. 
All data preprocessing procedures strictly follow the standard pipeline established by MedSAM \cite{ma2024nc_medsam}.
Four metrics are used: \textbf{1)} Dice measuring the overlap between the prediction and ground truth; \textbf{2)} NSD (Normalized Surface Dice) measuring the agreement between two surfaces within a specified tolerance; \textbf{3)} HD (Hausdorff Distance 95) quantifying the boundary error by computing the 95th percentile of the Hausdorff distance, lower better; \textbf{4)} clDice (centerline Dice) emphasizing the correctness of the centerline connectivity, especially suitable for vessel.

\noindent\textbf{Baselines.} 
We compare our method against three categories: 
\textbf{1)} Supervised learning methods \cite{isensee2021_NM_nnunet} directly trained on the target dataset; 
\textbf{2)} Foundation models \cite{isensee2025_arxiv_nninteractive, ma2025_medsam2}, where prompts are sampled from the ground truth mask during inference to guide the segmentation;
\textbf{3)} In-context learning methods \cite{wang2023_CVPR_Seggpt, butoi2023_ICCV_universeg, rakic2024cvpr_tyche, gao2025_CVPR_Iris} that do not require training or fine-tuning on the target dataset, relying on general-purpose pretrained models or inference-time adaptation.
For promptable foundation-model baselines, we follow the standard evaluation protocol and sample point prompts from the query ground-truth masks. This provides an optimistic oracle localization signal that is unavailable in deployment. In contrast, AtlasSegFM derives its prompts automatically from only the support atlas and the unlabeled query image through atlas-query registration. 

\noindent\textbf{Implementation details.}
Our method consists of two learnable components, which are test-time optimized: \textbf{1)} Atlas-registration module. We employ the Recursive Deformable Pyramid (RDP) network \cite{wang2024tmi_rdp}, which utilizes a dual-stream encoder and a pyramid decoder to recursively predict deformation fields from coarse to fine. It is optimized between support and query images using the Normalized Cross Correlation (NCC) loss with a learning rate of 1e-4 for 300 iterations. 
\textbf{2)} Adaptive fusion module. The fusion estimator $g(\cdot)$ is optimized at test time using a cycle-transformation Dice loss. Specifically, after support-to-query registration, the warped support mask is used as the atlas-derived prompt for the frozen foundation model, and the fused prediction is produced in the query space.
We then warp the fused prediction back to the support space and compute the Dice loss against the available support label. No query ground-truth mask is used during this optimization, and this is not a direct comparison between the support label and itself because the prediction has undergone deformation, foundation-model refinement, adaptive fusion, and inverse warping.
$g(\cdot)$ is updated for 100 iterations with a learning rate of $1e^{-5}$, while the foundation model remains frozen. 
For vascular segmentation, a recent foundation model vesselFM \cite{wittmann_2025CVPR_vesselfm} is adopted, whereas nnInteractive \cite{isensee2025_arxiv_nninteractive} is applied to other scenarios using mask-based prompts.

\subsection{Comparisons with State-of-the-art}
\begin{table}[t]
\fontsize{6.5pt}{9pt}\selectfont
\setlength{\tabcolsep}{2.58mm}
\setlength{\aboverulesep}{0.4ex}
\setlength{\belowrulesep}{0.4ex}
\setlength{\belowcaptionskip}{2mm}
\centering
\caption{Comparisons of organs-at-risk segmentation performance on the HaN-Seg and SegRap datasets, and pelvic bone segmentation performance on the Pengwin dataset.}
\begin{tabular}{l|cc|cc|cc}
\toprule
\multirow{2}{*}{\textbf{Methods}} & \multicolumn{2}{c|}{\textbf{HaN-Seg}} & \multicolumn{2}{c|}{\textbf{SegRap}} & \multicolumn{2}{c}{\textbf{Pengwin}} \\
\multicolumn{1}{c|}{} & \multicolumn{1}{c}{\textbf{Dice}$^\uparrow$} & \multicolumn{1}{c|}{\textbf{NSD}$^\uparrow$} & \multicolumn{1}{c}{\textbf{Dice}$^\uparrow$} & \multicolumn{1}{c|}{\textbf{NSD}$^\uparrow$} &  \multicolumn{1}{c}{\textbf{Dice}$^\uparrow$} & \multicolumn{1}{c}{\textbf{NSD}$^\uparrow$} \\
\hline
{\fontsize{3pt}{9pt}\selectfont nnUNet (supervised upper bound) \cite{isensee2021_NM_nnunet}} & 75.10 & 82.48 & 80.20 & 78.18 & 98.90 & 99.10\\
\hline
\rowcolor{gray!20} \multicolumn{7}{l}{$\blacktriangledown$ \textit{Foundation Models (1 or 5 clicks per test image)}} \\
\hline
MedSAM2-5 (2025) \cite{ma2025_medsam2} & 28.68 & 35.45 & 26.34 & 41.88 & 92.67 & 94.76 \\
nnInteractive-1 (2025) \cite{isensee2025_arxiv_nninteractive} & 42.42 & 52.08 & 45.67 & 56.84 & 72.09 & 75.29\\
nnInteractive-5 (2025) \cite{isensee2025_arxiv_nninteractive} & \textbf{55.29} & \textbf{66.24} & \textbf{50.02} & \textbf{69.32} & \textbf{93.63} & \textbf{96.00} \\
\hline
\rowcolor{gray!20} \multicolumn{7}{l}{$\blacktriangledown$ \textit{In-context Models (One support each context)}} \\
\hline
Register-based$^{1}$ {(TMI 2024)} \cite{wang2024tmi_rdp} & 48.72 & 67.96 & 44.50 & 66.95 & 82.75 & 85.25 \\
SegGPT (ICCV 2023) \cite{wang2023_CVPR_Seggpt} & 24.52 & 31.26 & 27.31 & 42.65 & 34.20 & 37.52 \\
UniverSeg (ICCV 2023) \cite{butoi2023_ICCV_universeg} & 35.49 & 45.33 & 39.28 & 56.86 & 56.10 & 63.03 \\
Tyche (CVPR 2024) \cite{rakic2024cvpr_tyche} & 33.99 & 43.10 & 37.45 & 54.22 & 42.74 & 46.90\\
Iris (CVPR 2025) \cite{gao2025_CVPR_Iris}  & 37.57 & 48.28 & 42.73 & 60.07 & 68.14 & 71.75 \\
\rowcolor{gray!20} Ours (MedSAM2-based) & 52.61 & 65.74 & 49.45 & 69.19 & 94.86 & 96.74 \\
\rowcolor{gray!20} Ours (nnInteractive-based) & \textbf{63.71} & \textbf{77.80}  & \textbf{69.19} & \textbf{87.39} & \textbf{95.50} & \textbf{96.96} \\
\bottomrule
\specialrule{0pt}{0pt}{2pt}
\multicolumn{7}{l}{${^1}$Using RDP \cite{wang2024tmi_rdp} to estimate the deformation field from the support to the query and apply } \\
\multicolumn{7}{l}{\hspace{4.5pt}it to the mask of support to obtain the mask of query. }
\end{tabular}
\label{tab:exp_abdmr_abdct}
\end{table}

\noindent\textbf{Results on organs-at-risk and pelvic bone datasets.}
Table \ref{tab:exp_abdmr_abdct} summarizes the Dice and NSD score comparisons across different methods on the HaN-Seg \cite{podobnik2023mp_hanseg}, SegRap \cite{luo2025MIA_segrap2023}, and Pengwin \cite{liu2023miccai_pengwin} datasets, focusing on the segmentation of organs-at-risk and pelvic bones. Our method (nnInteractive-based) achieves the highest mean Dice scores among all in-context models across all datasets, with 63.71\% on HaN-Seg, 69.19\% on SegRap, and 95.50\% on Pengwin, outperforming all baseline methods. 
Our approach achieves substantial performance gains over recent ICL methods, improving the mean Dice score on HaN-Seg by 26.14\% compared to Iris \cite{gao2025_CVPR_Iris} and 29.72\% compared to Tyche \cite{rakic2024cvpr_tyche}, highlighting its robust generalization and adaptability across diverse scenarios. 
Even compared to supervised learning methods, our approach achieves competitive performance, delivering segmentation quality comparable to nnUNet \cite{isensee2021_NM_nnunet} (e.g., 95.50\% vs. 98.90\% on Pengwin), which relies on full supervision and target-specific training. This highlights the effectiveness of our method in achieving high-quality segmentation without the need for extensive labeled data.

\begin{figure*}[t!]
\centering
\includegraphics[width=\linewidth]{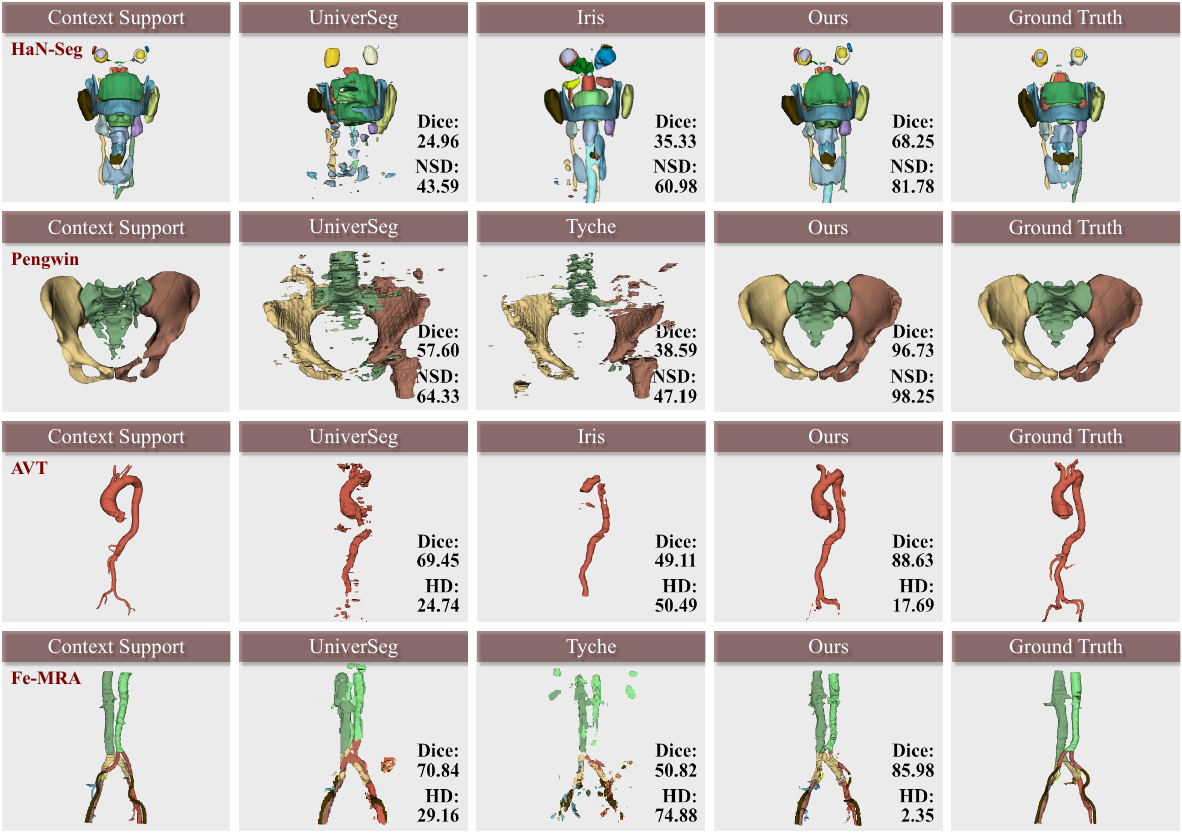}
\caption{Visual comparisons against recent one-shot learning methods with open-source implementations (i.e., UniverSeg \cite{butoi2023_ICCV_universeg}, Tyche \cite{rakic2024cvpr_tyche}, and Iris \cite{gao2025_CVPR_Iris}) across four datasets. Each color corresponds to a context, and all contexts are combined in the visualization.}
\label{fig:vis-all}
\vspace{0pt}
\end{figure*}

Qualitative results in Fig.~\ref{fig:vis-all} (first two rows) further validate our method’s capability to produce segmentations with clear boundaries and fine structural details. 
In contrast, others often struggle with incomplete or inconsistent outputs due to limited structural priors or contextual information.

\noindent\textbf{Results on brain datasets.} 
Table \ref{tab:brain} reports segmentation performance on the BrainRT dataset, where our method achieves the highest Dice and NSD scores across all metrics for both tasks.
For whole-brain segmentation, while most methods achieve reasonable performance, our approach stands out by delivering the highest accuracy with consistently smooth and anatomically precise segmentation, achieving a Dice score of 91.24\% and an HD of 6.78.
For the highly challenging organs-at-risk segmentation (e.g., optic nerves and brainstem), our method achieves a Dice score of 77.07\% and an NSD of 77.55\%, significantly outperforming all baselines by a massive margin. 
Accurate segmentation of these structures is crucial in radiotherapy to minimize severe radiation-induced damage; however, even the state-of-the-art Iris method struggles with unseen structures, underscoring the importance and difficulty of this task. 
As shown in Fig.~\ref{fig:vis-brain}, foundation models (e.g., nnInteractive 1/5 clicks) perform poorly or even fail on organs-at-risk segmentation due to the intricate structures and small size of these regions, which demand finer detail preservation and better contextual understanding.

\begin{table}[t]
\fontsize{6.5pt}{9pt}\selectfont
\setlength{\tabcolsep}{3.38mm}
\setlength{\aboverulesep}{0.4ex}
\setlength{\belowrulesep}{0.4ex}
\setlength{\belowcaptionskip}{2mm}
\centering
\caption{Comparisons of the brain-structure and organs-at-risk segmentation performance on the BrainRT dataset.}
\begin{tabular}{l|ccc|ccc}
\toprule
\multirow{2}{*}{\textbf{Methods}} & \multicolumn{6}{c}{~~\textit{Whole-brain}~~$\leftarrow$~~\textbf{BrainRT}~~$\rightarrow$~~\textit{Organs-at-risk}}\\
& \textbf{Dice}$^\uparrow$ & \textbf{NSD}$^\uparrow$ & \textbf{HD}$^\downarrow$ & \multicolumn{1}{c}{\textbf{Dice}$^\uparrow$} & \textbf{NSD}$^\uparrow$ & \textbf{HD}$^\downarrow$ \\ 
\hline
\rowcolor{gray!20} \multicolumn{7}{l}{$\blacktriangledown$ \textit{Foundation Models (1 or 5 clicks per test image)}} \\ \hline
nnInteractive-1 (2025) \cite{isensee2025_arxiv_nninteractive} & 0.62 & 1.82 & 104.04 &30.65 & 24.38& 90.01 \\  
nnInteractive-5 (2025) \cite{isensee2025_arxiv_nninteractive} & \textbf{61.63} & \textbf{40.31} & \textbf{27.62} & \textbf{39.09} &  \textbf{26.42} & \textbf{59.63} \\
\hline
\rowcolor{gray!20} \multicolumn{7}{l}{$\blacktriangledown$ \textit{In-context Models (One support each context)}} \\  \hline
SegGPT (ICCV 2023) \cite{wang2023_CVPR_Seggpt} &50.60 & 58.13 & 19.38 & 2.61 & 3.60 & 84.73 \\ 
UniverSeg (ICCV 2023) \cite{butoi2023_ICCV_universeg} & 83.32  & 32.44  & 15.40 &0.39 &  3.83& 32.59 \\
Tyche (CVPR 2024) \cite{rakic2024cvpr_tyche} &88.96 & 47.10 & 14.69 &4.37 & 10.03& 38.76\\
Iris (CVPR 2025) \cite{gao2025_CVPR_Iris} & 90.28 & 59.47& 8.56 & 10.45& 19.95& 44.52\\
\rowcolor{gray!20} Ours (nnInteractive-based) & \textbf{91.24} & \textbf{63.08} & \textbf{6.78} & \textbf{77.07} & \textbf{77.55} & \textbf{5.17} \\
\bottomrule
\end{tabular}
\label{tab:brain}
\vspace{0pt}
\end{table}

\begin{table}[t]
\fontsize{6.5pt}{9pt}\selectfont
\setlength{\tabcolsep}{2.95mm}
\setlength{\aboverulesep}{0.4ex}
\setlength{\belowrulesep}{0.4ex}
\setlength{\belowcaptionskip}{2mm}
\centering
\caption{Comparisons of the vessel segmentation on the AVT and Fe-MRA datasets.}
\begin{tabular}{l|ccc|ccc}
\toprule
\multirow{2}{*}{\textbf{Methods}} & \multicolumn{3}{c|}{\textbf{AVT}}        & \multicolumn{3}{c}{\textbf{Fe-MRA}}  \\
& \textbf{Dice$^\uparrow$}  & \textbf{clDice}$^\uparrow$ & \textbf{HD}$^\downarrow$ & \textbf{Dice}$^\uparrow$ & \textbf{clDice}$^\uparrow$ & \textbf{HD}$^\downarrow$ \\
\hline
\rowcolor{gray!20} \multicolumn{7}{l}{$\blacktriangledown$ \textit{Foundation Models (1 or 5 clicks per test image)}} \\
\hline
vesselFM (CVPR 2025) \cite{wittmann_2025CVPR_vesselfm} & 28.19 & 7.11   & 274.02 & \textbf{60.31} & \textbf{41.74}   & 124.34 \\
nnInteractive-1 (2025) \cite{isensee2025_arxiv_nninteractive} & 63.31 & 53.27  & 65.79  & 43.69 & 23.12   & 182.87 \\
nnInteractive-5 (2025) \cite{isensee2025_arxiv_nninteractive} & \textbf{83.38} & \textbf{70.52}  & \textbf{43.69}  & 49.36 & 34.00   & \textbf{110.01} \\ \hline
\rowcolor{gray!20} \multicolumn{7}{l}{$\blacktriangledown$ \textit{In-context Models (One support each context)}} \\  \hline
SegGPT (ICCV 2023) \cite{wang2023_CVPR_Seggpt} & 68.73 & 48.78 & 45.05 & 0.31 & 0.69 & 133.15 \\
UniverSeg (ICCV 2023) \cite{butoi2023_ICCV_universeg} & 39.71 & 31.89 & 51.12 & 69.71 & 53.20 & 39.61 \\
Tyche (CVPR 2024) \cite{rakic2024cvpr_tyche} & 53.24 & 42.76 & 70.44 & 50.21 & 59.64 & 79.28 \\
Iris (CVPR 2025) \cite{gao2025_CVPR_Iris} & 45.44 & 41.93 & 58.53 & 28.35 & 13.66 & 38.88    \\ 
\rowcolor{gray!20}Ours (vesselFM-based)  & \textbf{81.34} & \textbf{72.04} & \textbf{30.84} & \textbf{84.42} & \textbf{82.99} & \textbf{3.00} \\ 
\bottomrule
\end{tabular}
\label{tab:vessel}
\vspace{0pt}
\end{table}

\noindent\textbf{Results on vessel datasets.}
In Table~\ref{tab:vessel}, we evaluate on vessel segmentation, a challenging task due to intricate geometries and thin structures. Our method achieves the highest Dice scores (81.34\% on AVT and 84.42\% on Fe-MRA) and clDice scores (72.07\% on AVT and 82.99\% on Fe-MRA), while significantly outperforming other methods in HD (30.84 on AVT and 3.00 on Fe-MRA). These results demonstrate its capability to accurately capture complex vascular structures and maintain smooth, anatomically consistent surfaces.

As shown in Fig.~\ref{fig:vis-all} (last 2 rows), our method reconstructs the full topology of the aortic vessel tree on AVT and clearly delineates both arterial and venous structures on Fe-MRA, outperforming other methods that struggle with incomplete outputs. 
This highlights our method's ability to produce detailed, anatomically consistent segmentations.

\subsection{Ablation Studies}\label{sec:ablation}
Table \ref{tab:ablation} shows the ablation studies, where we evaluate the impact of different modules in our framework on the Pengwin, HaN-Seg, and Fe-MRA datasets. This analysis highlights the contributions of atlas-based registration, pre-registration strategies, foundation model prompts, and the fusion module. These datasets are selected to cover three representative clinical scenarios with different anatomical properties and potential failure modes: relatively rigid bony structures in Pengwin, deformable head-and-neck organs at risk in HaN-Seg, and thin, topology-sensitive vascular structures in Fe-MRA.

\begin{table}[t]
\fontsize{6.5pt}{9pt}\selectfont
\setlength{\tabcolsep}{0.95mm}
\setlength{\aboverulesep}{0.4ex}
\setlength{\belowrulesep}{0.4ex}
\setlength{\belowcaptionskip}{2mm}
\begin{minipage}[t]{0.58\linewidth}
\caption{Ablation with incremental addition of modules on the Pengwin and Fe-MRA datasets.}
\centering
\begin{tabular}{l|ccc}
\toprule
\textbf{Module Settings} & \textbf{Pengwin} & \textbf{HaN-Seg} & \textbf{Fe-MRA} \\ 
\hline
Baseline \#1 (Atlas) & 63.24 & 37.32 & 35.73 \\
~~+ Rigid pre-registration & 70.55 & 41.89 & 66.43 \\
~~+ Affine pre-registration & 82.75 & 48.72 & 81.64 \\
Baseline \#2 (FM) & ~~72.09$^\star$ & ~~42.42$^\star$ & 55.28 \\
~~+ Prompts from atlas & 94.56 & 61.74 & ~~N/A$^\dag$ \\
~~+ Fusion (Final) & \textbf{95.50} & \textbf{63.71} & \textbf{84.42} \\
\bottomrule
\specialrule{0pt}{0pt}{2pt}
\multicolumn{4}{l}{${^\star}$One point prompt is provided for nnInteractive.} \\
\multicolumn{4}{l}{${^\dag}$The foundation model vesselFM is not promptable.}
\end{tabular}
\label{tab:ablation}
\end{minipage}
~~~
\begin{minipage}[t]{0.388\linewidth}
\setlength{\tabcolsep}{1.4mm}
\setlength{\aboverulesep}{0.4ex}
\setlength{\belowrulesep}{0.4ex}
\setlength{\belowcaptionskip}{2mm}
\caption{Inference time (min per image) and learnable parameters.}
\centering
\begin{tabular}{l|c|c}
\toprule
\multirow{2}{*}{\textbf{Methods}} & \textbf{Inference} & \textbf{Learnable} \\
& \textbf{Time} & \textbf{Param.} \\
\hline
SegGPT & 18.1\, & 1.4G \\
UniverSeg & 2.2 & 1.2M \\
Tyche & 2.1 & 1.2M \\
\rowcolor{gray!20} Ours (Total) & 1.8 & 8.6M \\
\rowcolor{gray!20} ~~- Regist. & 1.5 & 8.5M \\
\rowcolor{gray!20} ~~- FM & 0.01 & -  \\
\rowcolor{gray!20} ~~- Fusion & 0.3 & 0.1M \\
\bottomrule
\end{tabular}
\vspace{0pt}
\label{tab:time}
\end{minipage}
\end{table}

\noindent\textbf{Atlas-FM registration.} The performance of Baseline \#1, which uses only atlas-based registration, sets the starting point for analysis. By simply aligning between atlas and query, the Dice scores are 63.24\% on Pengwin and 35.73\% on Fe-MRA. Pre-registration also improves the performance, with the rigid one raising Dice scores to 70.55\% on Pengwin and 66.43\% on Fe-MRA. Further incorporating affine pre-registration to better account for anatomical variations boosts performance to 82.75\% and 81.64\% on Pengwin and Fe-MRA.

\noindent\textbf{Foundation models.} Baseline \#2 evaluates the independent performance of the foundation model. For Pengwin, when we prompt nnInteractive with  one point per image, the Dice score is 72.09\%. On Fe-MRA, vesselFM achieves a Dice score of 55.28\%, leveraging its pre-trained capabilities for fine-grained segmentation. By providing atlas-derived prompts to the foundation model, the performance on Pengwin increases to 94.56\%, showcasing the utility of structural priors in guiding the foundation model.

\noindent\textbf{Atlas-FM fusion.} 
Atlas registration provides accurate global structure, whereas foundation models demonstrate better local details.
As listed in the last row, our adaptive fusion module combines the registered atlas prediction and the foundation model prediction, achieving Dice scores of 95.50\% on Pengwin and 84.42\% on Fe-MRA, surpassing either prediction alone and demonstrating the benefit of fusing their complementary strengths.


\noindent\textbf{Runtime and parameter efficiency.}
Table~\ref{tab:time} measures the computational cost on NVIDIA 4090 GPU using a ($256 \times 256 \times 240$) image from the BrainRT organs-at-risk task. Unlike slice-by-slice methods such as SegGPT, Tyche, and UniverSeg, our method operates directly on 3D images, enabling faster inference. It takes 1.8 minutes per image, compared to 18.1 minutes for SegGPT and 2.1 minutes for Tyche. The runtime mainly consists of atlas-based registration (1.5 minutes) and prediction fusion (0.3 minutes), whereas the foundation model inference takes 0.01 minutes. With only 8.6M learnable parameters and a single support-query pair for test-time adaptation, our method is lightweight, practical, and demonstrates relatively higher efficiency.

\begin{figure*}[t!]
\centering
\includegraphics[width=\linewidth]{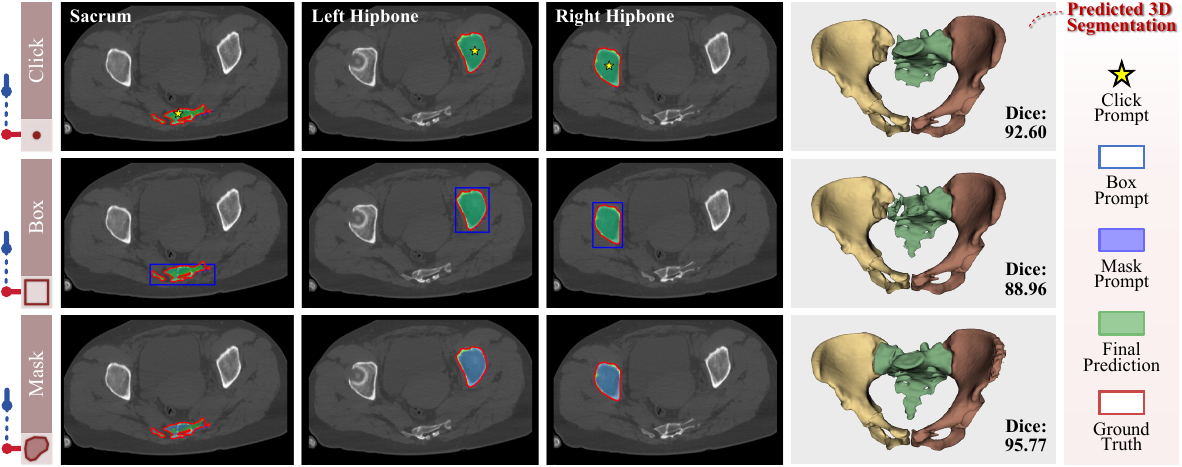}
\caption{Click, box, and mask prompts provided by our atlas guidance. Our method supports different foundation models/prompt types and achieves high accuracy across them, with mask
prompts performing best (visualized with nnInteractive).}
\label{fig:exp-prompt}
\vspace{-0pt}
\end{figure*}

\noindent\textbf{Different prompting types.} 
We analyze click/box/mask prompts for nnInteractive in Fig.~\ref{fig:exp-prompt}. On the Pengwin case, atlas-derived prompts yield 92.60 Dice (click), 88.96 Dice (box), and 95.77 Dice (mask), highlighting that dense mask prompts provide the most informative context when initialization is coarse.

\begin{figure*}[t!]
\centering
\includegraphics[width=\linewidth]{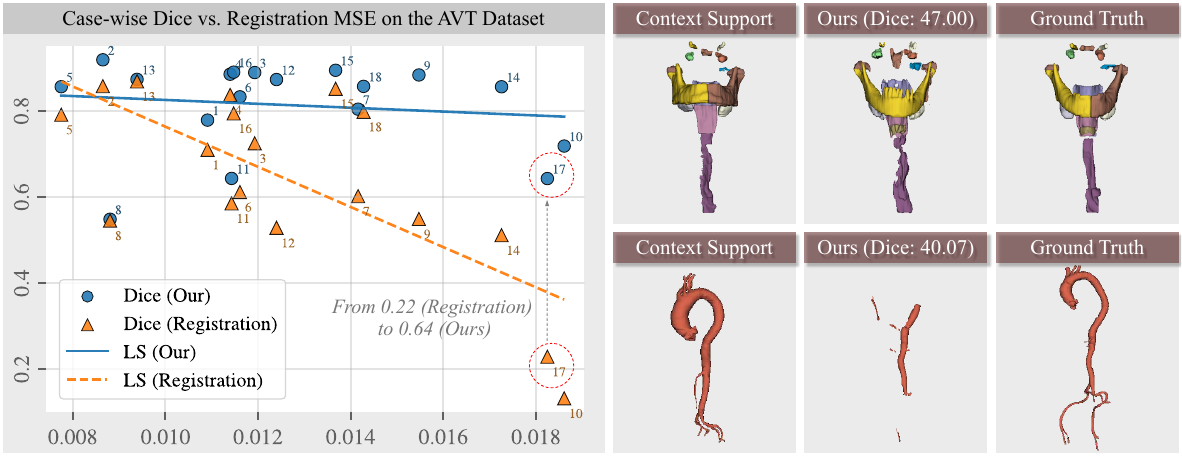}
\caption{Left: Comparison of the performance degradation under imperfect alignment. Right: Failure cases of our method. Top-right (from SegRap) erroneously merges the left and right mandibles. Bottom-right (from AVT) fails to fully recover the aortic arch. }
\label{fig:failure}
\vspace{-0pt}
\end{figure*}


\noindent\textbf{Sensitivity to registration quality.}
We further analyze how segmentation performance changes with registration quality.
As shown in Fig.~\ref{fig:failure} (left), Dice of registration degrades sharply as MSE increases, whereas our method shows a flatter degradation trend.
In the highlighted case, our method improves Dice from 0.22 to 0.64 after foundation-model refinement and adaptive fusion.
This suggests that our method does not simply inherit the warped atlas and can partially compensate for imperfect alignment.

\noindent\textbf{Failure cases.} 
Despite its strengths, our method encounters challenges in certain difficult cases. As shown in Fig.~\ref{fig:failure} (right), it erroneously merges the left and right mandibles in the SegRap dataset and misses critical regions in the AVT. The former arises because the left and right mandibles are artificially separated during annotation without a clear anatomical boundary, while nnInteractive tends to segment the mandible as a single continuous structure. The latter occurs when the support and query images differ substantially, making it difficult for the matching network to regress accurate correspondences.

%% file: sec/5_conclusion.tex
\section{Conclusion}
\label{sec:conclusion}
In this paper, we introduced AtlasSegFM, a simple yet effective medical image segmentation framework. By leveraging the globally consistent structure provided by the atlas and the representation power of foundation models, AtlasSegFM fuses their complementary strengths and achieves competitive accuracy.
It offers the following advantages:
\textbf{1)} It fully exploits off-the-shelf foundation models, enabling context-specific customization with a single atlas and no additional training data.
\textbf{2)} Experiments across six datasets demonstrate an average Dice of 80.35\%, substantially outperforming both uncustomized foundation models and existing in-context learning methods.
\textbf{3)} In rare contexts underrepresented in foundation model pretraining, such as organs-at-risk, our method achieves a Dice of 73.60\% without training on similar cases, demonstrating strong out-of-distribution robustness for complex, delicate structures.

\noindent\textbf{Limitations and future direction.}
While effective for organ segmentation, our method is less suitable for lesion segmentation due to the dependence on structural consistency between support and query images, which focal lesions often lack. It also faces challenges with patients whose anatomy differs markedly from the atlas. Future work will explore multi-atlas strategies for greater robustness and refine prompt generation for more context-aware guidance.